\documentclass[10pt,twocolumn,letterpaper]{article}

\usepackage{iccv}
\usepackage{times}
\usepackage{epsfig}
\usepackage{graphicx}
\usepackage{amsmath}
\usepackage{amssymb}

\usepackage[pagebackref=true,breaklinks=true,letterpaper=true,colorlinks,bookmarks=false]{hyperref}

\usepackage[capitalize]{cleveref}
\crefname{section}{Sec.}{Secs.}
\Crefname{section}{Section}{Sections}
\Crefname{table}{Table}{Tables}
\crefname{table}{Tab.}{Tabs.}

\usepackage{titletoc}

\usepackage{amsmath,amsfonts,bm}

\def\eqref#1{equation~\ref{#1}}

\def\1{\bm{1}}

\def\rva{{\mathbf{a}}}

\def\rvf{{\mathbf{f}}}
\def\rvg{{\mathbf{g}}}

\def\rvq{{\mathbf{q}}}

\def\rvv{{\mathbf{v}}}

\DeclareMathAlphabet{\mathsfit}{\encodingdefault}{\sfdefault}{m}{sl}
\SetMathAlphabet{\mathsfit}{bold}{\encodingdefault}{\sfdefault}{bx}{n}

\newcommand{\softmax}{\mathrm{softmax}}

\DeclareMathOperator*{\argmax}{arg\,max}

\usepackage{url}
\usepackage[utf8]{inputenc} 
\usepackage[T1]{fontenc}    
\usepackage{xcolor}         
\definecolor{mydarkblue}{rgb}{0,0.08,0.45}
\usepackage{hyperref}       
\hypersetup{ %
    pdftitle={},
    pdfauthor={},
    pdfsubject={},
    pdfkeywords={},
    pdfborder=0 0 0,
    pdfpagemode=UseNone,
    colorlinks=true,
    linkcolor=black,
    citecolor=mydarkblue,
    filecolor=mydarkblue,
    urlcolor=mydarkblue,
    pdfview=FitH}
\usepackage{url}            
\usepackage{booktabs}       
\usepackage{amsfonts}       
\usepackage{nicefrac}       
\usepackage{microtype}      

\usepackage{multirow}
\usepackage{subcaption}
\usepackage{bigstrut}
\usepackage{graphicx}
\usepackage{amsmath}
\usepackage{amssymb}
\usepackage{algorithm}
\usepackage{algorithmicx}
\usepackage{algpseudocode}
\usepackage{tabu}
\usepackage{siunitx}
\usepackage{adjustbox}
\usepackage{subcaption}
\usepackage{siunitx}
\usepackage{makecell}
\usepackage{colortbl}
\usepackage{color}
\usepackage{multirow}
\usepackage{blindtext}
\usepackage{comment}
\usepackage{bm}
\usepackage{bbm}
\usepackage{booktabs}
\usepackage{wrapfig}
\usepackage{array} 
\definecolor{Gray}{gray}{0.95}
\definecolor{Cyan}{rgb}{0.88,1,1}
\definecolor{LightCyan}{rgb}{0.92,1,1}
\definecolor{DarkCyan}{rgb}{0.82,1,1}
\newcommand{\ourmeos}{\textbf{\texttt{Tem-adapter}} }
\newcommand{\traf}{SUTD-TrafficQA } 
\newcommand{\msr}{MSR-VTT-MC }

\usepackage{wrapfig}

\newcommand{\tabstyle}[1]{
  \setlength{\tabcolsep}{#1}
  \centering
  \small
}

\iccvfinalcopy 

\def\iccvPaperID{3975} 

\ificcvfinal\fi

\begin{document}

\title{Tem-adapter: Adapting Image-Text Pretraining for Video Question Answer} 
\author{Guangyi Chen$^{1,2}$\footnotemark[1]~, Xiao Liu$^{5}$\footnotemark[1]~, Guangrun Wang$^{4}$, Kun Zhang$^{1,2}$, Philip H.S. Torr$^{4}$, \\ Xiao-Ping Zhang$^{3,6}$, Yansong Tang$^{3}$\footnotemark[2]~\\
$^{1}$Carnegie Mellon University, Pittsburgh PA, USA \\
$^{2}$Mohamed bin Zayed University of Artificial Intelligence, Abu Dhabi, UAE \\
$^{3}$Shenzhen International Graduate School, Tsinghua University, China \\
$^{4}$University of Oxford, UK \\
$^{5}$Eindhoven University of Technology, NL \\
$^{6}$Toronto Metropolitan University, Canada\\
}


\ificcvfinal\thispagestyle{empty}\fi

\vspace{-2em}
\twocolumn[{
    \renewcommand\twocolumn[1][]{#1}
    \maketitle
    \begin{center}
        \includegraphics[width=1.0\linewidth]{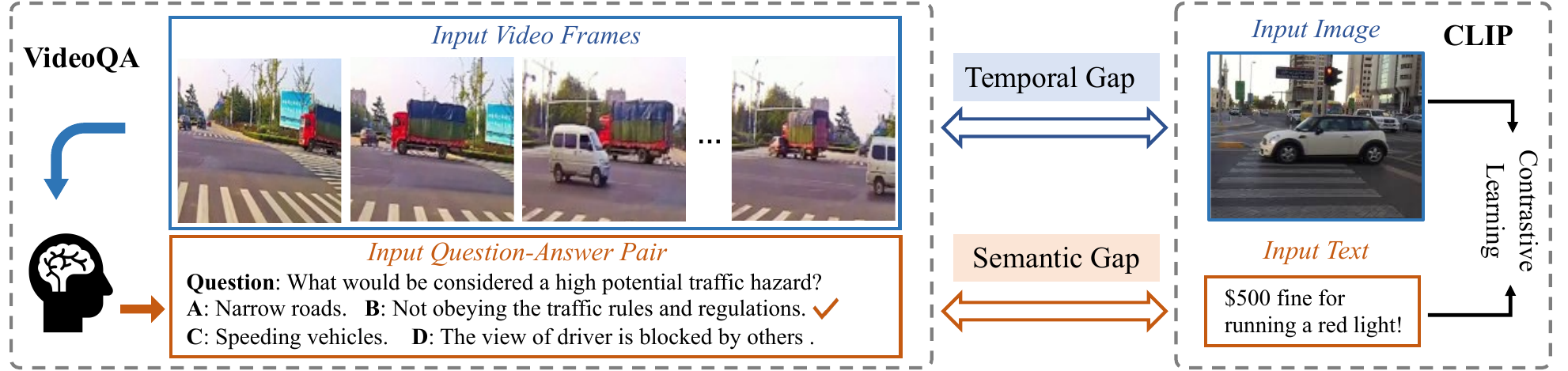}
        \vspace{-0.5cm}
        \captionof{figure}{Illustration of the domain gaps between the pre-trained CLIP and the downstream VideoQA. CLIP is trained to align visual and textual domains, while VideoQA requires understanding the temporal dynamics and complex semantics of videos.}
        \label{fig:teaser}
    \end{center}
}]
\ificcvfinal\thispagestyle{empty}\fi

\begin{abstract}
\vspace{-0.1cm}
Video-language pre-trained models have shown remarkable success in guiding video question-answering (VideoQA) tasks. However, due to the length of video sequences, training large-scale video-based models incurs considerably higher costs than training image-based ones. This motivates us to leverage the knowledge from image-based pre-training, despite the obvious gaps between image and video domains.  
To bridge these gaps, in this paper, we propose \textbf{\texttt{Tem-Adapter}}, which enables the learning of temporal dynamics and complex semantics by a visual Temporal Aligner and a textual Semantic Aligner.
Unlike conventional pre-trained knowledge adaptation methods that only concentrate on the downstream task objective, the Temporal Aligner introduces an extra language-guided autoregressive task aimed at facilitating the learning of temporal dependencies, with the objective of predicting future states based on historical clues and language guidance that describes event progression. 
Besides, to reduce the semantic gap and adapt the textual representation for better event description, we introduce a Semantic Aligner that first designs a template to fuse question and answer pairs as event descriptions and then learns a Transformer decoder with the whole video sequence as guidance for refinement. 
We evaluate \textbf{\texttt{Tem-Adapter}} and different pre-train transferring methods on two VideoQA benchmarks, and the significant performance improvement demonstrates the effectiveness of our method. \footnote[1]{Our code can be found at: \href{https://github.com/XLiu443/Tem-adapter}{https://github.com/XLiu443/Tem-adapter}}

\end{abstract}
\renewcommand{\thefootnote}{\fnsymbol{footnote}}
\footnotetext[1]{Equal contribution.}
\footnotetext[2]{Corresponding author.}

\vspace{-0.32cm}

\section{Introduction}
\label{sec:intro}

Video Question Answering (VideoQA) is a task that aims to answer natural language questions based on the information available in observed videos. It has attracted considerable attention recently due to its promise to develop interactive AI systems capable of communicating with the dynamic visual environment using natural language. Despite significant advancements in recent years, VideoQA remains a challenging problem that requires models to comprehensively understand and dynamically align the semantics of both the visual world and natural language.

Recently, a variety of methods \cite{yang2022frozenbilm,yang2021just,zellers2022merlot,clipbert,Videoclip,seo2021look} have demonstrated impressive results in utilizing the large-scale vision-language pre-trained (VLP) model to enhance downstream VideoQA tasks. These methods pre-train models by aligning the video and language domains and then apply them to VideoQA tasks via fine-tuning or even zero-shot learning. Nevertheless, pre-training large-scale models necessitate a large number of video-text pairs, \textit{e.g.,} more than 10 million videos, and entails expensive computational costs. This motivates us to explore cheaper and lighter alternative pre-trained models.


Using image-VLP models is one potential option as they also align the semantics of vision and language domains. Compared to video-based models, image-based models offer two significant advantages. Firstly, training image-based models is less expensive when considering an equal number of data pairs. Secondly, image-VLP models have made significant progress in recent years, with some of them widely available and used, such as CLIP~\cite{clip}. However, as depicted in Figure~\ref{fig:teaser}, domain gaps persist between image-based pre-training and VideoQA tasks since image-based models are unable to learn the temporal dynamics and corresponding complex event semantics of the visual world. 



To address the challenge posed by the domain gaps, this paper proposes \ourmeos, an adapter network that leverages the interaction between visual and textual modalities to learn temporal dynamics and complex semantics. Unlike existing methods that directly adapt visual and textual features with the objective of the downstream tasks, such as cross-modal matching in VideoQA, \textbf{\texttt{Tem-Adapter}} introduces an additional language-guided autoregressive task to facilitate learning of temporal dynamics, which enables visual representations to predict future states based on historical information and language guidance.
Besides, we introduce cross-modal interactions to further reduce the semantic gap and refine textual representation.


Our approach consists of a visual Temporal Aligner and a textual Semantic Aligner. The Temporal Aligner uses a Transformer encoder to learn the temporal relations and refine visual features. To optimize it, we build an auto-regression model by a Transformer decoder to generate the future state with historical information. We also incorporate the language information for visual refinement by adding textual embeddings as the condition memory of the Transformer decoder.
This auto-regression model is supervised by a reconstruction loss with ground truth future frames to encourage the learning of temporal dependencies. 
In addition to the visual branch, we introduce a Semantic Aligner for better event description by reducing the semantic gap between the original texts crawled from the net and question-answer pairs in VideoQA.  First, we design templates to fuse textual questions and answers to generate declarative sentences to reduce the domain gap between training and downstream languages. Then, a Transformer decoder is employed to refine textual embeddings by incorporating the entire video sequence as a memory condition for video-text interactions.

We conduct experiments on two public VideoQA benchmarks including SUTD-TrafficQA~\cite{trafficqa} and MSR-VTT-MC~\cite{msrvtt}, and evaluate different categories of methods that transfer the pre-trained knowledge to downstream tasks, such as other adapter models~\cite{clip-adapter}, prompt learning~\cite{coop,vpt}, and fully or partial finetuning. 
The significant improvement demonstrates the effectiveness of \ourmeos to adapt the image-language pre-trained model for VideoQA tasks.

\section{Related Work}
\label{sec:related}

\textbf{Video Question Answering.} 
Compared with Image-based Question  Answering, Video QA is more challenging due to the complex temporal dependency, which requires models to dynamically align the visual observation and natural language. Some previous methods~\cite{tapaswi2016movieqa,lifeqa,mun2017marioqa,ye2017video,kim2017deepstory,song2018explore,xue2018better,kim2020modality,kim2020dense,chadha2020iperceive,colas2019tutorialvqa,choi2021dramaqa,dang2021object,li2019beyond,zha2019spatiotemporal,zhuang2020multichannel,fan2019heterogeneous,gao2018motion,lei2019tvqa+,garcia2020knowit,huang2020location,jiang2020divide,jiang2020reasoning,park2021bridge,sadhu2021video,nextqa} learn the models with the given manually annotated datasets. Recently, more and more methods~\cite{yang2022frozenbilm,yang2021just,zellers2022merlot,clipbert,Videoclip,seo2021look,kim2021self} propose to utilize the representation ability of large-scale pre-trained vision-language models to solve the VideoQA task.  For example, VQA-T~\cite{yang2021just} pre-trains the vision language model with randomly sampled video-question-answer triplets in the HowTo100M~\cite{howto100m} dataset. 
BiLM~\cite{yang2022frozenbilm} pre-trains frozen bidirectional language models with some adapter layers on the WebVid10M~\cite{bain2021frozen} dataset and achieve excellent zero-shot VideoQA performance on the downstream tasks.
However, pretraining the large-scale model is expensive, which motivates us to find a cheaper alternative manner.
In this paper, we propose to use existing image-language pre-trained models, such as CLIP~\cite{clip}, to provide the alignment between the visual and textual domains. The most related method is ATP~\cite{atp}, which also applies CLIP as the pre-trained model to extract frame and language embeddings, and learns an atemporal probe to select the most discriminative frames to match the video and QA pairs. Different from ATP selecting frames to transfer the video to images, we propose \ourmeos to learn the temporal dynamics and reduce the gap between the pre-trained image domain and downstream video domain.

\begin{figure}[t]
\begin{center}
\centerline{\includegraphics[width=1.0\linewidth]{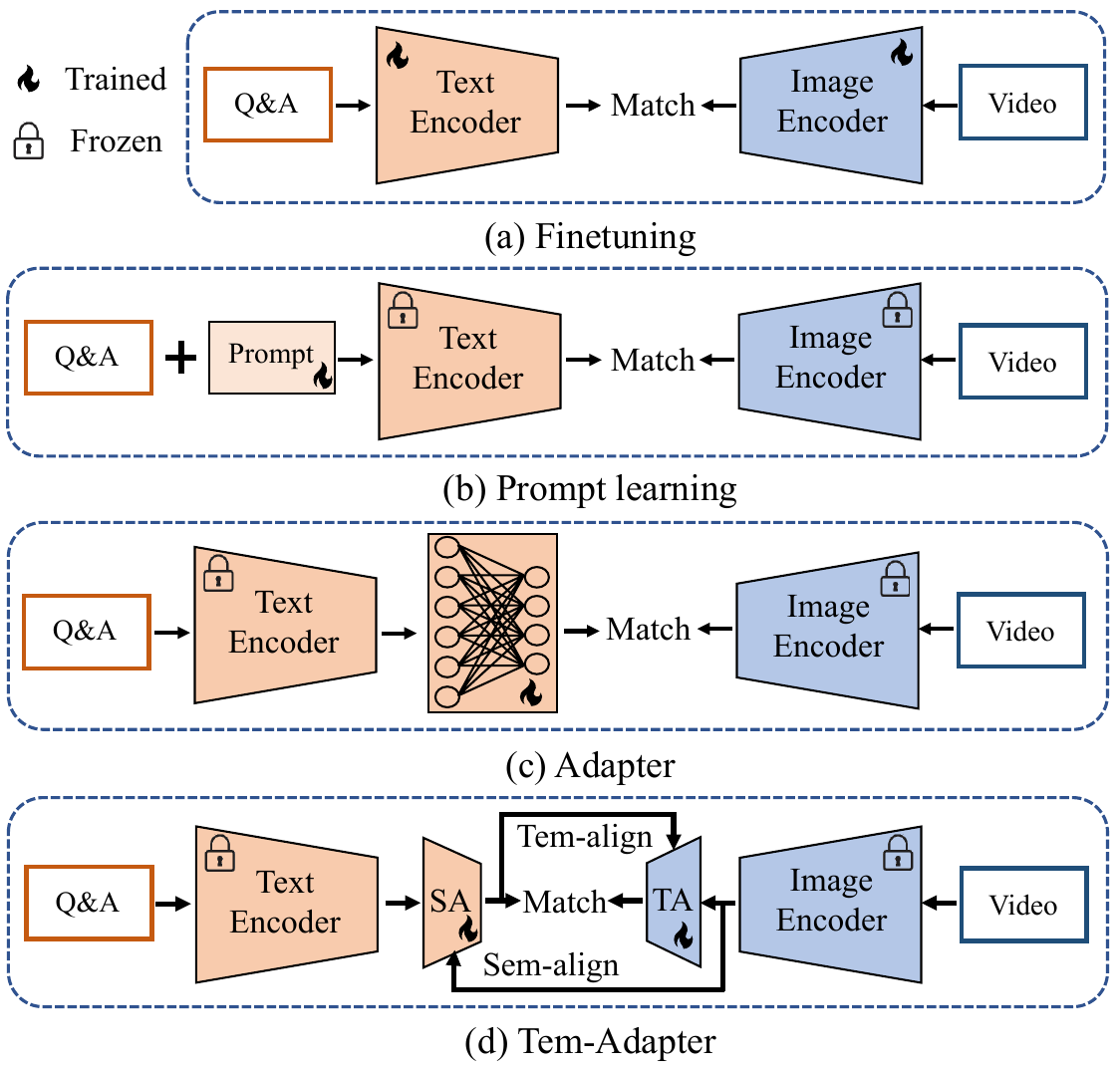}}
\caption{\textbf{Various methods for pre-trained knowledge adaptation in VideoQA
.} 
(a) Fine-tuning updates pre-trained parameters with downstream training. (b) Prompt learning freezes the pre-trained model and introduces learnable textual tokens for adaptation. (c) Adapter-based methods add new adapter layers to transfer knowledge. (d) Our approach, \ourmeos, leverages the visual-textual interactions to learn temporal dynamics to reduce the gap between image-based pre-training and video-based tasks.
}
\label{fig:top}
\end{center}
\vspace{-0.9cm}
\end{figure}

\textbf{Vision Language Pre-training.} 
The goal of VLP is to help downstream tasks by pre-training the model on large-scale vision-text pairs~\cite{clip,blip,align}.  It always crawls a huge amount of vision-text pairs from the web and trains the model in a self-supervised manner, such as reconstruction~\cite{visualbert,vln-bert,meter,vilt}, contrastive matching~\cite{clip,align,mural}, or the combination of both two~\cite{li2021align,vlmo,kamath2021mdetr}. 
Different from image-VLP, video-VLP has more diverse proxy tasks, such as frame/language mask~\cite{videobert,actbert}, video-language matching~\cite{miech2020end}, video/language generation~\cite{xu2021vlm,luo2020univl}, and frame/language order prediction~\cite{lei2021understanding,hero}.
Recently, VLP models have shown great potential for other downstream applications, such as zero-shot and few-shot visual recognition~\cite{clip,clip-adapter,coop,cocoop,vtclip}, segmentation and detection ~\cite{rao2021denseclip,zhou2021denseclip}, image generation~\cite{nichol2021glide,ramesh2022hierarchical,patashnik2021styleclip}, and action understanding~\cite{wang2021actionclip,tevet2022motionclip}.
In this paper, we use the CLIP model as our pre-trained model.

\begin{figure*}[t]
\begin{center}
\centerline{\includegraphics[width=0.95\linewidth]{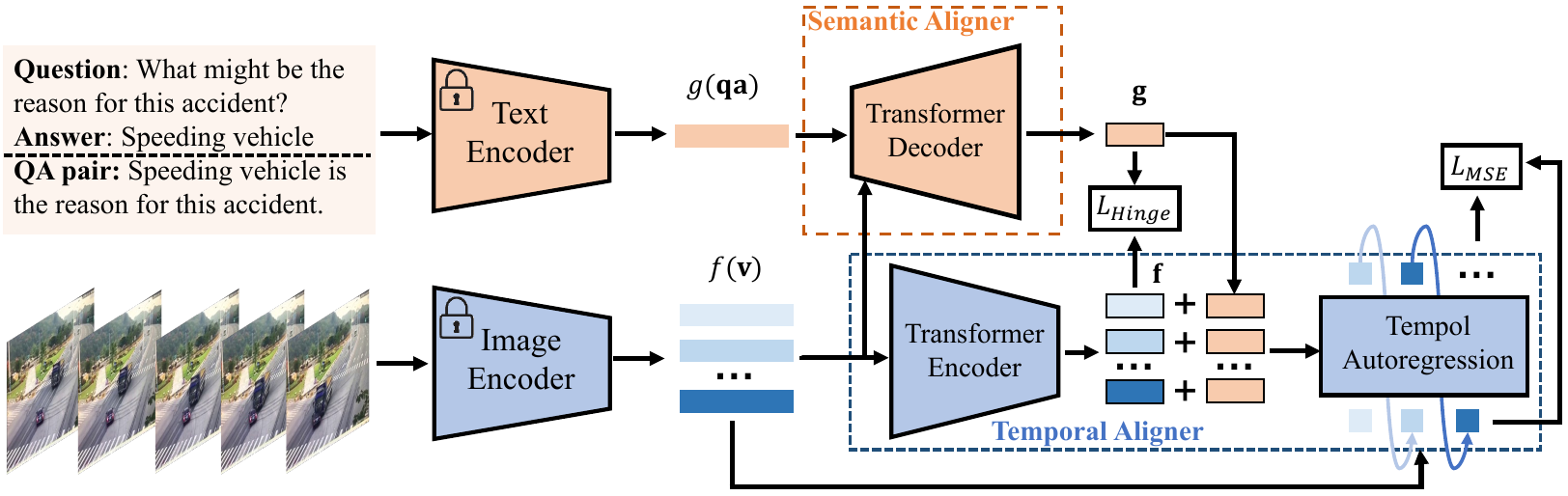}}
\vspace{-0.3cm}
\caption{The overall framework of \ourmeos. Our \ourmeos consists of two modules, Semantic Aligner (orange) and Temporal Aligner (blue). Semantic Aligner takes the text embedding of template-based QA pair as the input, and applies a Transformer decoder to incorporate cross-modal interactions.
The Temporal Aligner introduces a language-guided autoregressive task to learn temporal dynamics. The training is guided by classification and reconstruction losses. }
\label{fig:main}
\end{center}
\vspace{-0.8cm}
\end{figure*}

\textbf{Pre-trained Knowledge Transfer.}
In light of the remarkable success of the large-scale pre-trained model, many methods 
have been proposed to solve the problem that how to efficiently adapt the pre-trained knowledge to the downstream tasks.
Early methods finetune all parameters of the pre-trained model with the downstream objectives~\cite{fastrcnn,peters2019tune}. However, with the model scale increasing, the cost of finetuning the full model is extremely expensive. Thus, an increasing number of methods focus on knowledge adaptation with light parameters. A straightforward way is to partially finetune the model parameters~\cite{nakamura2019revisiting}. Besides, prompt learning methods are proposed recently to reformulate the downstream tasks as the pre-trained tasks and freeze all model parameters~\cite{liu2021pre,petroni2019language,radford2019language,poerner2019bert}.
The previous prompt-based methods~\cite{petroni2019language,poerner2019bert} design the template of prompt using the prior of natural human language. Then, some methods~\cite{jiang2020can,autoprompt} propose to automatically optimize the prompt template, and ~\cite{li2021prefix,tsimpoukelli2021multimodal,liu2021gpt,coop,plot} further learn the prompts in the continuous embedding space where the discrete word constraint is relaxed.
In addition, some methods freeze the model and train some extra adapter layers and also achieve the supersizing performance~\cite{clip-adapter,tip-adapter,bain2021frozen}.
As shown in Figure~\ref{fig:top}, we illustrate the difference between \ourmeos and other categories of methods transferring the image-based VLP model to the VideoQA task. 
Compared with other pre-train adaptation methods, our \ourmeos proposes an autoregression task to reduce the temporal gap (tem-align) and introduce the cross-domain interactions (sem-align) to reduce the semantic gap between the pre-training and downstream tasks.



\section{Approach}
\label{sec: method}

In this section, we first briefly introduce the VideoQA task and then present our approach, \ourmeos, which introduces the cross-modal interactions to learn temporal dynamics and complex semantics. Finally, we give the details of our implementation.

\subsection{Problem Definition }
The goal of VideoQA is to find an answer $\hat{\rva}$ from the answer pool, given the conditions of question $\rvq$ and observed video data $\rvv$ as below:
{
\setlength{\abovedisplayskip}{5pt}
\setlength{\belowdisplayskip}{5pt}
\begin{equation} \label{eq:video}
\hat{\rva} = \argmax_{\rva \in \mathcal{A}} \mathcal{S} (\rva|\rvv,\rvq),
\end{equation}
}
where $S $ is a match function to generate the score of each answer candidate given the question and video content.


\subsection{Tem-Adapter}
Here, we propose \ourmeos to learn temporal dynamics and complex semantics by the interactions of visual and textual domains to adapt pre-trained Image-VLP models to downstream VideoQA tasks.
Figure~\ref{fig:main} shows an overview of \textbf{\texttt{Tem-Adapter}}'s framework. In particular, \ourmeos consists of two components, including a visual Temporal Aligner and a textual Semantic Aligner. Details for each component are as follows.


\textbf{Temporal Aligner.}
The goal of the Temporal Aligner is to reduce the gap between image-based pre-training and video-based downstream tasks. To achieve this goal, we introduce an auxiliary language-guided autoregressive task to facilitate the learning of temporal dynamics.

Specifically, given a video $\rvv$, we apply the frozen CLIP~\cite{clip} image encoder $f$ to extract the feature of each 
frame as $f(\rvv)=\{f(\rvv_t)|_{t=1}^{T}\} \in \mathcal{R}^{T \times C}$, where $T$ is the frame number of videos and $C$ is the feature dimension of CLIP. Then we refine these features with our Temporal Aligner model which is built by a transformer encoder $f_e$ with parameters $\phi$ to learn the temporal relations of frames. The refined features can be represented as:
\begin{equation} \label{eq:autoregression_e}
\rvf=f_e(f(\rvv);\phi),
\end{equation}
which are with the same dimension $\mathcal{R}^{T \times C}$.

To guide the Temporal Aligner to learn temporal dynamics, we further build an autoregressive sequential model, following GPT~\cite{radford2018improving}, which generates the feature of the current frame with previous frames and guidance from textual information that describes event progression.
Mathematically, it can be formulated as:
{
\setlength{\abovedisplayskip}{5pt}
\setlength{\belowdisplayskip}{5pt}
\begin{equation} \label{eq:autoregression_d}
\hat{f}(\rvv_i) =   f_d(f(\rvv_{1:i-1})|\rvf,\rvg; \psi),
\end{equation}
}
where $f(\rvv_{1:i-1})$ is the input of the transformer decoder, which is achieved by applying an attention mask to focus on the history clues. $\rvf,\rvg$ are used as the guidance information, where $\rvf$ denotes latent visual embeddings obtained by Temporal Aligner and $\rvg$ (see \eqref{eq:transdecoder}) is the textual embeddings describing temporal event. $\rvf$ and $\rvg$ are first fused by adding and then serve as the memory key and value of the Transformer decoder. We apply a reconstruction loss to encourage the learning of temporal dependencies which is formulated as the distance between the results of the auto-regression model $\hat{f}(\rvv_i)$ and the ground-truth $f(\rvv_i)$:
{
\setlength{\abovedisplayskip}{5pt}
\setlength{\belowdisplayskip}{5pt}
\begin{equation} \label{eq:autoregression_loss}
L_{MSE} = \sum_{i=1}^{n} ||f(\rvv_i) - \hat{f}(\rvv_i)||_2^2.
\end{equation}
}

\textbf{Semantic Aligner.}
In addition to the visual branch, we introduce a Semantic Aligner to adapt the textual representation for better event description and reduce the semantic gaps. 
We first design templates to align the question-answer pairs in VideoQA to declarative sentences, since the training data of CLIP is crawled from the web which always describes the image in a declarative manner.
Then, we apply a Transformer decoder with the video sequence as the memory for video-text interactions.

Given a question $\rvq$ and a corresponding answer $\rva$, we first apply the grammar parsing on the question $\rvq$ to obtain the subject-verb-object structure, and then we modify the order of the sentence to obtain a declarative sentence and leave the position of the interrogative pronouns as a mask `[]'. Finally, we fill in the default `[]' with the answer to update the QA pair as a refined event description. For instance, the question is \textit{`Which might be the reason for this accident?’}, and the answer is  \textit{‘Improper lane change'}. We first obtain a declarative sentence with the default as \textit{`[ ] is the reason for this accident’}, and then we can fill in the answer to get \textit{`Speeding vehicle is the reason for this accident.'} In addition, for the general question, we first change it to a declarative sentence and then convert the sentence to negative form if the answer is No.

With the refined sentence $\rvq\rva$ as the input, the frozen text encoder $g$ of the CLIP model outputs the textual feature as $g(\rvq\rva) \in \mathcal{R}^C$. 
To further refine the textual feature using the video content, we apply a transformer decoder with the visual feature sequence as the memory key and value and add a residual link to avoid overfitting. 
Formally, it can be formulated as follows:
{
\setlength{\abovedisplayskip}{5pt}
\setlength{\belowdisplayskip}{5pt}
\begin{equation} \label{eq:transdecoder}
\rvg = g(\rvq\rva)+ \lambda g_d(g(\rvq\rva), f(\rvv); \theta) ,
\end{equation}
}
where $\rvg$ denotes the learned textual embedding, $\lambda$ is a parameter of the residual link, and $\theta$ is the parameters of the transformer decoder $g_d$. Both $\lambda$ and $\theta$ can be learned automatically in the training process.
With the cross-attention module in the transformer decoder, we encourage the model to find key information from the video to refine the textual embedding for better event description.

\textbf{Cross-modal Matching.}
Given the refined visual and textual embeddings $\rvf$ and $\rvg$,  we instantiate the score function in \eqref{eq:video} by cosine similarity as:
\begin{equation} \label{eq:softmax}
S(\rva|\rvv,\rvq) = \softmax(\bar{\rvf}^T\rvg),
\end{equation}
where $\bar{\rvf}$ is video representation using the average pool on all frames, $\rvg$ is the language embedding with given answer question pair ($\rva$, $\rvq$). Then, we can obtain the answer to the question with the highest matching score.

Then we apply a standard multi-choice classification Hinge loss~\cite{jang2017tgif} integrated with the auto-regression loss for training. The  overall loss function can be described as:
{
\setlength{\abovedisplayskip}{5pt}
\setlength{\belowdisplayskip}{5pt}
\begin{equation} \label{eq:loss}
L = L_{Hinge}(S(\rva|\rvv,\rvq),y) + \gamma L_{MSE},
\end{equation}
}
where $\gamma$ is a ratio to balance two loss functions, and $y$ is the ground truth label of the answers. Please note that our \ourmeos consists of 3 learnable modules including the textual decoder $\theta$, the visual encoder $\phi$, and the visual decoder $\psi$. The Hinge loss is applied to optimize $\theta$ and $\phi$, while the auto-regression loss $L_{MSE}$ is used to optimize all three modules. In the inference, only textual decoder $\theta$ and visual encoder $\phi$ are used.

\subsection{Implementation Details}
In this subsection, we provide more implementation details of our approach for better reproduction.

\textbf{Network Architecture.}
Figure \ref{fig:main} shows the framework of \ourmeos. In our experiments, we use the frozen model CLIP (ViT-B32) as the text and image encoders. 
The Semantic Aligner is a one-layer Transformer decoder with 16 heads. There is a linear layer before the Transformer decoder which maps text embeddings and image embeddings to a lower dimension (128). Another linear projection layer is added after the Transformer decoder to map the reduced dimension  back to 512. The Temporal Aligner is a light transformer encoder-decoder architecture. Both the encoder and decoder contain only one layer. We set the number of heads to 16 in the experiments.

\textbf{Inference Details.}
In the inference (given 1 question, 1 video, and $k$ answers), we first apply the template to fuse the question and $k$ answers to obtain $k$ QA pairs (e.g. $k=4$). Then we feed the QA pairs and video into CLIP textual and visual encoders to obtain the initial features. After that, we apply the learned semantic and temporal aligners to refine these features. Please note that the temporal auto-regression module is not used in the inference.  Finally, we obtain 1 video feature with $k$ QA features and apply the cosine similarity as well as softmax to select the answer.

\textbf{Hyper-parameters.}
We follow the same video frames sampling method with prior methods \cite{trafficqa}. We uniformly select 8 frames over a long video sequence and then pick consecutive 16 frames which are centered on those 8 frames. Overall 128 frames are selected for each video. We also tried another sampling method: uniformly extracting 128 or 256 frames on a video sequence, and the final results do not support the sampling method very well. In the loss function, we set $\gamma$ as 100 to encourage the model to learn the temporal dynamics. And the learnable parameter $\lambda$ in the Semantic Aligner is set to be the same dimension (512) as the output embeddings of the CLIP model .

\textbf{Training Details.}
We use the pre-trained CLIP model to extract text embeddings and image embeddings. The Adam \cite{kingma2014adam} optimizer is utilized to optimize the training process. We set the initial learning rate to be $1e-4$ with a batch size of 128.
And we let the learning rate decay with a factor of 0.5 for every 10 epochs. Our model is trained for 50 epochs on a single Tesla V100 GPU and the training takes 8 hours to finish. Our model is implemented in PyTorch 1.8.0.

\section{Experiments}
In this section, we first benchmark our \ourmeos and different categories of methods that transfer the pre-trained model into downstream tasks. Then we provide ablation studies  to investigate the effect of each component.

\subsection{Datasets \& Experimental Protocols}
\textbf{SUTD-TrafficQA.}
\traf focuses on the question-answering task in the traffic situation, which requires the model to understand the traffic event and underlying causal relation. It consists of more than 10k videos with different traffic events and provides over 62,535 human-annotated QA pairs.
Among them, 56,460 QA pairs are used for training and the rest 6,075 QA pairs are used for testing.
This dataset highlights the cognitive capability of video event understanding models in complex traffic scenarios. It provides  6 challenging traffic-related reasoning tasks including ``Basic understanding'', ``Event forecasting'', ``Reverse reasoning'', ``Counterfactual inference'', ``Introspection'', ``Attribution''. 
All tasks are formulated as multiple-choice without limiting the number of candidate answers. 

\textbf{MSR-VTT-MC.} MSR-VTT is collected and released mainly for the text-to-video retrieval task. Following ~\cite{clipbert} and ~\cite{atp}, we use the multi-choice test setting to benchmark the VideoQA methods. 
The \msr test set consists of 2990 QA pairs. Each video corresponds to five answers, with the original caption as the correct answer.
We use the standard protocol with prior works in our experiments, where 
7010 train+val video sequences and 140,200 QA pairs are utilized for training. Each video only contains one caption. We assign this caption as the correct answer for our question-answering task. To formulate four negative answers during training, we randomly select four captions from other captions that do not belong to the current video.


\begin{table*}
\caption{VideoQA accuracies of baseline methods on both \traf and \msr. \textbf{B}: ``Basic understanding'', \textbf{F}: ``Forecasting task'', \textbf{R}: ``Reverse Reasoning'', \textbf{C}: ``Counterfactual inference'', \textbf{I}: ``Introspection'', \textbf{A}: ``Attribution''. (C) and (A) denote strategies for adding prompts. (C) and (C$^*$) denote training prompts with/without adapter heads.}
\vspace{-0.4cm}
\linespread{3.0}
\renewcommand\arraystretch{1.1}
\renewcommand\tabcolsep{8pt}
\begin{center}
\newcolumntype{g}{>{\columncolor{Gray}}c}
\newcolumntype{y}{>{\columncolor{LightCyan}}c}
\newcolumntype{d}{>{\columncolor{DarkCyan}}c}
\begin{tabular}{l| c c c c c c g| g}
\hline
\hline
\multirow{2}{*}{\textbf{Methods}} &\multicolumn{7}{c}{\textbf{SUTD-TrafficQA}}& \multicolumn{1}{|c}{\textbf{MSR-VTT-MC}} \\ 
\cline{2-9}
~ &\multicolumn{1}{c}{\textbf{B}} & \multicolumn{1}{c}{\textbf{F}} &  \multicolumn{1}{c}{\textbf{R}} &  \multicolumn{1}{c}{\textbf{C}} & 
\multicolumn{1}{c}{\textbf{I}} &  \multicolumn{1}{c}{\textbf{A}} &
\multicolumn{1}{c}{\textbf{Avg}}& \multicolumn{1}{|c}{\textbf{Acc}}
\\
\hline
Unsupervised CLIP~\cite{clip} & 25.6 & 20.1 &34.0   & 30.8 & 22.8 & 28.8 &  26.5 &74.1  \\ 
CLIP~\cite{clip} + Template & 31.8 & 36.0 & 29.9 & 71.8   & 22.1 &  33.4 & 32.3   & 74.1  \\ 
Totally finetuning & 39.8 & 35.1 & 46.6 & 45.6 & 37.2  & 40.5  & 40.3 & 89.0   \\ 
Partially finetuning  & 41.6 & 37.8 &  44.6  & 50.0  & 33.1 & 41.7 & 41.7 & 88.5 \\ 
LORA~\cite{hu2021lora} & 38.7 & 38.7 &  36.7  & 37.9  & 34.5 & 38.1 & 38.3 & 85.4 \\
CLIP-Adapter~\cite{clip-adapter} &  35.8 &  32.0 &  35.4 &  42.3 & 33.1   &32.1  &34.8   &83.4  \\ 
Multi-layer Adapter~\cite{clip-adapter} &  30.5 &  26.6& 26.5 & 38.5 & 28.3  & 25.8 & 29.1  & 90.7 \\ 
Prompt learning (C)~\cite{coop} & 42.4  &  32.4  &   45.2  & 55.5  &  40.7 & 43.6   & 42.9 & 89.0  \\ 
Prompt learning (C$^*$)~\cite{coop} &  40.3 & 33.2   &  41.0   &  46.5  & 34.9  &   38.4 &39.7  & 90.8   \\
Prompt learning (A)~\cite{vpt} &  41.7  &  31.5 &  40.1 & 48.4  & 33.1  &  41.4 & 41.1 &  88.0 \\ \hline
\ourmeos &  \textbf{46.0} &   \textbf{36.5} &  \textbf{44.6} &   \textbf{55.0} &   \textbf{34.5}  &   \textbf{47.7} &  \textbf{46.0} & \textbf{94.3}   \\ 
\hline\hline
\end{tabular}
\end{center}
\label{table:base}
\vspace{-0.8cm}
\end{table*}

\begin{table}
\caption{Ablation studies on the \traf and \msr datasets. Tp, SA, TA, and Ar denote Template, Semantic Aligner, Temporal Aligner, and Auto-regression, respectively. No ablation studies (``$\backslash$'') concentrate on Tp, as it is not compatible with the \msr dataset.
}
\vspace{-0.4cm}
\linespread{3.0}
\renewcommand\arraystretch{1.1}
\renewcommand\tabcolsep{2pt}
\begin{center}
\newcolumntype{g}{>{\columncolor{Gray}}c}
\newcolumntype{y}{>{\columncolor{LightCyan}}c}
\newcolumntype{d}{>{\columncolor{DarkCyan}}c}
\begin{tabular}{l| c | c}
\hline
\hline
\multirow{2}{*}{\textbf{Methods}} &\multicolumn{1}{c}{\textbf{SUTD-TrafficQA}}& \multicolumn{1}{|c}{\textbf{MSR-VTT-MC}} \\ 
\cline{2-3}
~ 
&\multicolumn{1}{c}{\textbf{Avg}} &
\multicolumn{1}{|c}{\textbf{Acc}}
\\
\hline
CLIP-Adapter~\cite{clip-adapter}   &34.8 &83.4  \\ \hline
w/o Tp & 43.9 & $\backslash$   \\
w/o SA & 44.2 &91.8  \\ 
w/o TA  &  43.6 &91.6 \\ 
w/o Ar  &  44.9 &92.6\\ 
{w/o TA and SA}  &32.3& 74.1   \\ 
\hline
\ourmeos &  \textbf{46.0} & \textbf{94.3}  \\ 
\hline\hline
\end{tabular}
\end{center}
\label{table:ablation}
\vspace{-0.7cm}
\end{table}

\subsection{Baseline Methods}
To investigate different categories of methods that transfer the pre-trained model into downstream tasks, such as finetuning, prompt learning, and adapter. We compare \ourmeos with the other 10 baseline methods including Unsupervised CLIP~\cite{clip}, Unsupervised CLIP~\cite{clip} + Language template, Totally fine-tuning, Partially fine-tuning, LORA~\cite{hu2021lora}, CLIP-Adapter~\cite{clip-adapter}, Multi-layer CLIP-Adapter~\cite{clip-adapter}, Prompt learning (change words)~\cite{coop} (with/without using adapter heads), and Prompt learning (add words)~\cite{vpt}. The detailed descriptions are shown in the \textbf{supplementary materials}.

\subsection{Benchmarking against Baseline Methods}

In this subsection, we compare and discuss our approach’s performance against the above-mentioned baselines on both \traf and \msr datasets. The results of all methods on both datasets are summarized in Table~\ref{table:base}. We can draw conclusions as follows from the results.

\textbf{CLIP can obtain good zero-shot results.} With our language template, the clip model can achieve $32.3\%$ accuracy on the \traf dataset, which is almost close to the previous state-of-the-art performance ($37.1\%$). It shows the strong representation ability of the CLIP model.

\textbf{More parameters $\neq$ Better performance.} We have two groups of experiments that focus on the number of parameters, including totally/partially finetuning and single-layer/multi-layer Adapters. We obtain contradictory conclusions from different datasets. On SUTD-TrafficQA, the model with fewer parameters works better, while on MSR-VTT-MC, the model with more parameters has better performance. It is not surprising since the scale of training data is different. 
\msr has more training data and can support training more parameters. 
We also compare the learnable parameters of adapter-based methods in Table~\ref{tab:model_size}. \ourmeos achieve better performance with fewer parameters than Multi-layer Adapter. 
 Compared with the prompt-learning-based method,  our method needs more parameters, but its inference is not time-consuming.

\textbf{Different categories of methods are comparable.} Balancing the performance on both datasets, different categories of methods that transfer the image-based VLP model to VideoQA don't have clear performance differences. Adapter-based methods obtain poor results on SUTD-TrafficQA, but multi-layer Adapter works well on MSR-VTT-MC.  
Besides, by comparing the prompt learning methods with/without adapter heads,  we found that the prompt and adapter can be joint learning to obtain better performance. 

\textbf{Cross-modal interaction and temporal dynamics matter.}  Compared with CLIP-Adapter (both single-layer and multi-layer ones), our \ourmeos achieves more than $10\%$ improvement on SUTD-TrafficQA. The key difference between \ourmeos and other adapter-based methods is that we introduce temporal alignment to reduce the temporal gap and apply the cross-modal interaction to refine the representation and reduce the semantic gap. \ourmeos outperforms other methods by a large margin on both two datasets. For example, it obtains $3.1\%$ improvement than Prompt learning (C) on \traf and $3.6\%$ than Multi-layer Adapter on MSR-VTT-MC.





\begin{table}[t]
\caption{Comparison with the state-of-the-art methods on the \traf dataset. 
}
\vspace{-0.4cm}
\setlength{\tabcolsep}{10pt}
\begin{center}
\begin{tabular}{l | c | c}
\hline
\hline
\textbf{Methods} & \textbf{Accuracy} &  \textbf{Year} \\
\hline
\textit{Human}    &  95.4   & -- \\ 
\hline
VIS+LSTM \cite{ren2015exploring} & 29.9  & 2015\\
TVQA \cite{lei2018tvqa} & 35.2   &2018  \\ 
HCRN \cite{le2020hierarchical} & 36.5  &2020 \\
BERT-VQA \cite{yang2020bert} & 33.7 & 2020 \\ 
Eclipse~\cite{trafficqa}  & 37.1  & 2021\\ 
CLIP~\cite{clip}  & 32.3 & 2021 \\ 
CLIP+HCRN\cite{clip, le2020hierarchical}   & 41.9 & 2021\\
ATP~\cite{atp}  & 35.6 &2022 \\ 
\hline
\ourmeos & \textbf{46.0} \\ \hline
\hline
\end{tabular}
\end{center}
\label{table:baseline_comparison}
\vspace{-0.79cm}
\end{table}

\subsection{Comparison against State-of-the-Art Methods}
\textbf{SUTD-TrafficQA.}
On SUTD-TrafficQA, we compare our proposed method with state-of-the-art methods, including ~\cite{le2020hierarchical,trafficqa,ren2015exploring,lei2018tvqa,yang2020bert}. The results are summarized in Table~\ref{table:baseline_comparison}. We can observe that our \ourmeos outperforms other methods by a large margin. For example, compared with HCRN~\cite{le2020hierarchical}, \ourmeos achieves almost $10\%$ improvement. 
By using the CLIP model to extract visual features, HCRN+CLIP can obtain $41.9\%$ accuracy, which is better than using the original HCRN.
Besides, we show that the performance improvement is not only from the better representation ability of the CLIP model since our \ourmeos further improve HCRN+CLIP with $+4.1\%$ accuracy. 
 In addition, we also significantly outperform the ATP method, which also uses the CLIP model.
This significant performance improvement demonstrates the effectiveness of \ourmeos.

\textbf{MSR-VTT-MC.}
On MSR-VTT-MC, we compare \ourmeos with some recent state-of-the-art methods, including ActBERT\cite{actbert}, ClipBERT\cite{clipbert}, MERLOT \cite{zellers2021merlot}, VideoCLIP \cite{videobert}, CLIP~\cite{clip}, and ATP~\cite{atp}. Results are shown in Table~\ref{tbl:vid-qa:msrvttmc}. First, we can find that the CLIP model benefits the performance on the \msr dataset. All methods using the CLIP model, such as CLIP~\cite{clip}, ATP~\cite{atp}, and \ourmeos achieve good performance.  ATP~\cite{atp} learns an atemporal probe to select a frame from a video sequence to obtain the visual embedding and achieve good performance on MSR-VTT-MC. Compared with it, \ourmeos can further obtain $94.3\%$ accuracy and $1.1\%$ relative improvement.

\begin{table}[t]
\caption{Comparison with the state-of-the-art methods on the \msr dataset. 
}
\vspace{-0.4cm}
\linespread{3.0}
\renewcommand\arraystretch{1.1}
\setlength{\tabcolsep}{10pt}
\begin{center}
\begin{tabular}{l | c | c}
\hline
\hline
\textbf{Methods} & \textbf{Accuracy}  & \textbf{Year} \\
\hline
VideoCLIP \cite{videobert} & 92.1 &2019\\
ActBERT\cite{actbert} & 85.7 &2020\\
ClipBERT\cite{clipbert} & 88.2 &2021 \\
MERLOT \cite{zellers2021merlot} & 90.9 &2021\\
CLIP~\cite{clip}  & 74.1 &2021\\
ATP~\cite{atp} & 93.2 &2022\\
\hline
\ourmeos & \textbf{94.3} \\ 
\hline
\hline
\end{tabular}
\end{center}
\label{tbl:vid-qa:msrvttmc}
\vspace{-0.7cm}
\end{table}

\begin{figure*}[t]
\begin{center}
\centerline{\includegraphics[width=1\linewidth]{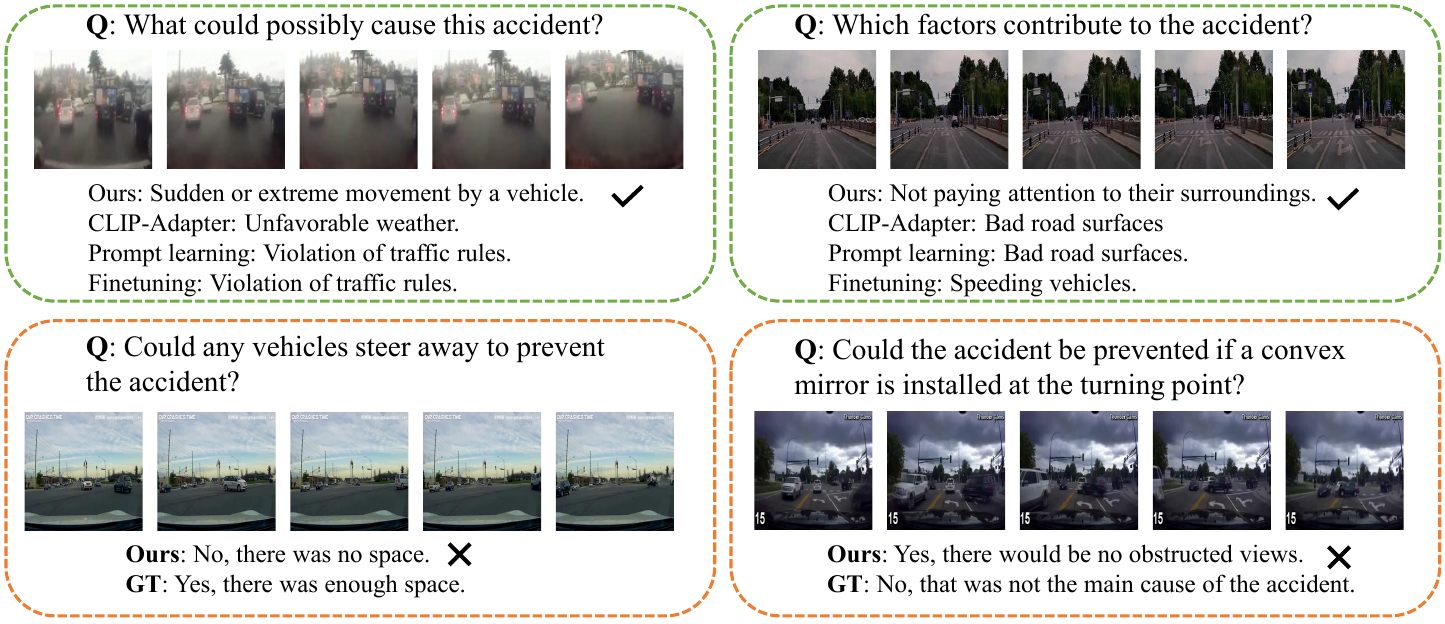}}
\caption{Visualization of some examples of the VideoQA task from the SUTD-TrafficQA dataset. The top row shows some positive examples, where \ourmeos learns temporal dynamics to understand the traffic event and give the correct answers. The bottom row shows some failed cases. Despite better matching between visual and textual domains, our approach fails on complex problems which require reasoning ability.   }
\label{fig:vis}
\end{center}
\vspace{-0.8cm}
\end{figure*}

\subsection{Ablation Studies}
We provide ablation studies to investigate the effect of each component. Four main modules in our \ourmeos are evaluated, including Temporal Aligner, Semantic Aligner, language template, and auto-regression.

\begin{table*}[t]
\renewcommand\tabcolsep{3pt}
\small
\begin{center}
\caption{
The comparisons of our method and other baseline methods on model parameters (M), inference time (ms/video), and accuracy in the SUTD-TrafficQA dataset.} 
\vspace{-0.2cm}
\begin{tabular}{c|ccccccc}
\hline
Metrics & LORA~\cite{hu2021lora} & Prompt learning (C)~\cite{coop} & Prompt learning (C$^*$)~\cite{coop} & CLIP-Adapter~\cite{clip-adapter}  & Multi-layer Adapter~\cite{clip-adapter} & Ours\\
\hline
Parameters & 1.44 & 3.55 & 0.77 & 1.05  &  4.85  &  3.62 \\
Inference time & 5.37 & 5.33 & 5.17 & 5.13 & 5.14 & 5.32 & \\
Accuracy & 38.3 & 42.9 & 39.7 & 34.8 & 29.1 & \textbf{46.0} & \\
\hline
\end{tabular}
\label{tab:model_size}
\end{center}
\vspace{-0.9cm}
\end{table*}


\begin{table}[t]
    \tabstyle{6pt}
    \small
    \caption{The comparison of reconstruction ability. We use peak signal-to-noise ratio (PSNR) 
 as the metric.}
    \label{tab:reconstrution}
     \vspace{-0.2cm}
	 \setlength{\tabcolsep}{8pt}
	\begin{tabular}{lccccc}
	 \toprule
    & \ourmeos & w/o $\rvg$ & w/o $\rvf$ and $\rvg$ \\
\midrule
 PSNR & 29.1 & 28.7 & 27.4 \\
    \bottomrule
    \end{tabular}
    \vspace{-0.5cm}
\end{table}

As shown in Table~\ref{table:ablation}, we provide the performance of different variants of \ourmeos and a baseline adapter method on each task in the \traf and \msr datasets.
The baseline method is CLIP-adapter, which also uses the language template and learns a linear adapter layer without cross-domain interaction.
We observe that all variants of \ourmeos can achieve significant improvement over the baseline method, which shows that the interaction across visual and textual is very important for video analysis. Besides, compared with \ourmeos, we can observe a clear performance drop when we remove each component, which demonstrates all of them are indispensable. When both temporal and semantic aligners are removed, the performance drops dramatically.

In addition, we also conducted an experiment to show whether the Temporal Aligner learns the temporal dynamic information. Table~\ref{tab:reconstrution} shows the reconstruction quality of our methods with/without language guidance ($\bm{g}$) and with/without Temporal Aligner ($\bm{f}$). We can observe that both the language guidance and Temporal Aligner assist in the learning of temporal dynamics.

Furthermore, we also provide some parameter analysis experiments about our network design. And evaluate \ourmeos on different image-text pre-train models. Please kindly refer to the \textbf{supplementary materials} for the detailed results and analysis.

\subsection{Visualization}
In this subsection, we provide some qualitative results to analyze our approach. As shown in Figure~\ref{fig:vis}, we show some visualization examples of VdieoQA on the \traf dataset. In the top raw, we plot two positive examples. We can observe that our \ourmeos learns the temporal dynamics with the cross-domain interactions to obtain a better understanding of video events. Besides, we also provide some failure cases to help explore the boundary of our approach. We find that \ourmeos can not work well for situations that need complex reasoning, especially for the underlying cause-effect relations. We leave it as a future work to enhance the reasoning ability of our model. Besides, we also provide more examples of our language templates in the \textbf{supplementary materials}.

\section{Conclusion}

In this paper, we proposed \ourmeos to adapt the pre-trained image-language model to the downstream VideoQA task. We jointly learn the Temporal Aligner and Semantic Aligner through cross-modal interactions. \ourmeos introduces a language-guided autoregressive task to guide the learning of temporal dependency and thus reduce the temporal gap between image-based pre-train and video-based QA tasks. Besides, we utilize a rule-based template and video-based interaction to refine textual representation and reduce semantic gaps. We validate the effectiveness of our approach on four VideoQA benchmarks.

\textbf{Broader Impacts and Limitations.} Understanding how the big model adaptation works on VideoQA is of great help in developing general foundation models and interactive AI. However, our work also has certain limitations. For example, since extracting video features online takes up too many computing resources, we extract their features offline and fix them. We leave it as future work to learn the adapter within the backbone network. 



\section*{{Acknowledgment}}
This work was supported in part by the National Natural Science Foundation of China under Grant 62206153, Young Elite Scientists Sponsorship Program by CAST (No. 2023QNRC002), by the Tsinghua Shenzhen International Graduate School-Shenzhen Pengrui Endowed Professorship Scheme of Shenzhen Pengrui Foundation, by the UKRI grant: Turing AI Fellowship EP/W002981/1, by the National Institutes of Health (NIH) under Contract R01HL159805, by the NSF-Convergence Accelerator Track-D award \#2134901, by a grant from Apple Inc., a grant from KDDI Research Inc, and generous gifts from Salesforce Inc., Microsoft Research, and Amazon Research. We would also like to thank the Royal Academy of Engineering and FiveAI.

{\small
\bibliographystyle{ieee_fullname}

}

\clearpage

\bigskip
{\noindent \large \bf {APPENDIX}}\\
\renewcommand{\thesection}{A-\arabic{section}}
\renewcommand{\thetable}{A-\arabic{table}}
\renewcommand{\thefigure}{A-\arabic{figure}}
\setcounter{table}{0}
\setcounter{figure}{0}
\setcounter{section}{0}

\section{Language Template}
In this section, we provide some examples of the language template utilized in the Vision-guided Language Adapter to see how the language template works. The steps of transferring a question and the corresponding answer to a declarative sentence are explained in Section 3.2.
Table \ref{tab:template-generation} shows examples of the 6 challenging traffic-related reasoning tasks including
“Basic understanding”, “Event forecasting”, “Reverse reasoning”,
“Counterfactual inference”, “Introspection”, “Attribution”. Table \ref{tab:template-generation-questiontype} provides examples with various question types including: "Where", "Why", "How", "How many", "What's", "Are there", "Did". 
It shows that all types of question-answer pairs can be transferred to declarative sentences with our language template.

\section{Details of Baseline Methods}
To investigate different categories of methods that transfer the pre-trained model into downstream tasks, such as finetuning, prompt learning, and adapter. We compare \ourmeos with the other 9 baseline methods in Sections 4.2 and 4.3 of the main submission. In this section, we provide a detailed description of these methods as follows:
\begin{itemize}
    \item Unsupervised CLIP~\cite{clip}:  The most direct manner to use the pre-trained clip model is the unsupervised manner, which uses image and text encoders to obtain the visual and textual features and match them with cosine distance, where the QA pair is connected as one sentence. 
    \item Unsupervised CLIP~\cite{clip} + Language template: Using a predefined template to transfer the QA pair into a declarative sentence, to reduce the language style gap between pre-train and downstream domains. We use this template for all the following baseline methods on SUTD-TrafficQA. Please note that \msr doesn't provide QA pairs but a caption, thus we don't use the template for this dataset.
    \item Totally finetuning: Totally finetuning denotes that we finetune all parameters of the CLIP model.
    \item LoRA~\cite{hu2021lora}: We add the LoRA module to the text encoder of the CLIP model. Each transformer layer of the encoder is adapted with LoRA. 
    \item Partially finetuning: Partial finetuning indicates that we only finetune a part of model parameters, such as the projection layers. 
    \item CLIP-Adapter~\cite{clip-adapter}: CLIP-Adapter adds a linear layer following the CLIP textual encoder and then freezes the encodes and learns this linear layer with downstream losses (classification). 
    \item Multi-layer CLIP-Adapter~\cite{clip-adapter}: To evaluate the effect of more parameters, we use a multi-layer perceptron as the adapter translation.
    \item Prompt learning (change words)~\cite{coop}: Given a sequence of tokens of the QA pair, it changes a part of tokens as the learnable parameters and learns them to better align video and texts. Note that we add the adapter heads in this method.
    \item Prompt learning (change words) without adapter heads~\cite{coop}: Similar to the previous Prompt learning (change words), it learns the parameters of tokens but the adapter heads are removed.
    \item Prompt learning (add words)~\cite{vpt}: Different changing word tokens as learnable parameters, an alternative is to add some learnable word tokens before the QA pair. The adapter heads are also added.
\end{itemize}


\section{Parameter Analysis}
In Section 4.5, we discussed the importance of each component of the \ourmeos, and it is observed that the performance drop if either the VL Adapter or the LV Adapter is removed. 
In this section, we further investigate hyperparameters' effects in the \ourmeos. We implemented experiments on the SUTD-TrafficQA dataset.
We tried different hyper-parameters of both the Semantic Aligner and the Temporal Aligner, including the latent dimension and the number of layers in the Semantic Aligner, the number of encoder layers, and the number of decoder layers in the Temporal Aligner.
Results are shown in Table \ref{table:param}. We set the latent dimension of the Semantic Aligner to 64 and 256. 
Also, the layer number of the Semantic Aligner is changed to 1 or 2. 
It is observed that the performance is similar to the best accuracy of our model, which indicates the robustness of the training model.
In addition, we adjusted the number of encoder layers and the number of decoder layers in the Temporal Aligner and obtained comparable results. 
The stable performance illustrates our model is robust enough under different hyper-parameter settings.

\section{Additional Visualizations}
Qualitative results are shown in Section 4.6 in our main submission. To better understand our method, we provide more examples in Figure \ref{fig:vis}.
Positive examples can support that our \ourmeos is able to learn  the temporal dependencies of videos with the language information and visual embeddings. 
An example is shown in the top right of Figure \ref{fig:vis}. It can be observed a small car is hit by a van when driving and the rear area is badly damaged. Our model is able to capture the temporal dynamics of the video and align the correct text information with the visual dependencies. 
In addition, Two more failure cases are included to show that our model can not behave well on samples that need complex reasoning. In the bottom right of Figure \ref{fig:vis},  the question and answer are closely related to the causes of the accident. That requires exploring interactions and relationships between the components of the video. We find that our model may fail under the complex causal reasoning and leave it as our further work.

\begin{table}[h]
\caption{Parameter analysis on the SUTD-TrafficQA dataset. \textbf{D} denotes the latent dimension of our Semantic Aligner; \textbf{LN} denotes the transformer layer number in Semantic Aligner; \textbf{ELN} and \textbf{DLN} respectively denote the layer numbers of the Encoder and Decoder in our Temporal Aligner. }
\vspace{-0.4cm}
\linespread{3.0}
\renewcommand\arraystretch{1.2}
\renewcommand\tabcolsep{8pt}
\begin{center}
\begin{tabular}{c c|c c|c}
\hline
\hline
\multicolumn{4}{c|}{\textbf{Parameters}} & \multirow{3}{*}{\textbf{Accuracy}} \\
\cline{1-4}
\multicolumn{2}{c}{\textbf{Semantic Aligner}}& \multicolumn{2}{|c|}{\textbf{Temporal Aligner}} \\ 
\cline{1-4}
\multicolumn{1}{c}{\textbf{D}} & \multicolumn{1}{c}{\textbf{LN}} & 
\multicolumn{1}{|c}{\textbf{ELN}} &  \multicolumn{1}{c|}{\textbf{DLN}} 
\\
\hline
 64 & 1   & 1   &  1 &45.3\\ 
64 & 2   & 1 & 1 &44.6 \\ 
256 & 1  &  1 & 1 & 45.1 \\ 
256 & 2  & 1 &  1 & 45.4\\ 
\cline{1-5}
128  & 1 & 1 & 2 & 45.6 \\
128 & 1  & 2 & 1   & 45.8 \\ 
128 & 1  & 2 & 2  & 45.8 \\ 
\cline{1-5}
128 &   1   & 1  &  1 & 
 \textbf{46.0}\\ 
\hline\hline
\end{tabular}
\end{center}
\label{table:param}
\vspace{-0.8cm}
\end{table}

\begin{figure*}[h]
\begin{center}
\centerline{\includegraphics[width=1.0\linewidth]{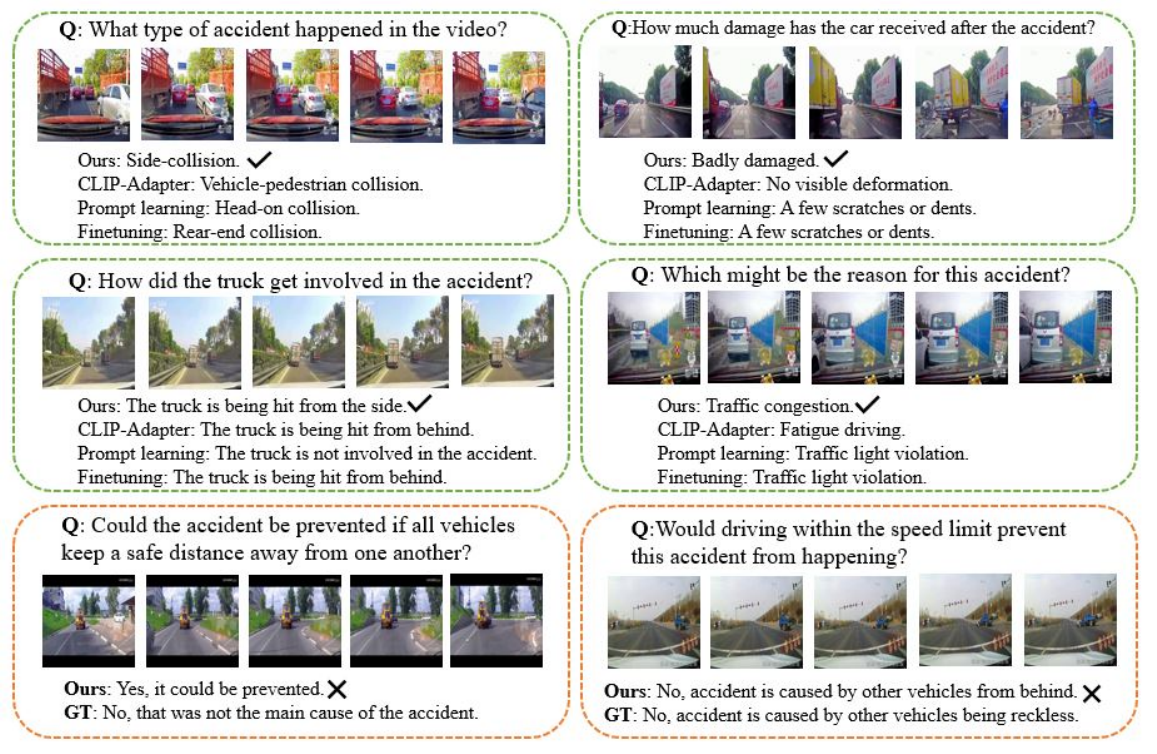}}
\caption{Visualization of more examples of the VideoQA task from the SUTD-TrafficQA dataset. The top two rows show four positive examples, in which \ourmeos learns temporal dependencies of videos to understand the traffic event. Correct candidates are selected by the \ourmeos.  
The bottom row includes two failed cases. Our model can not behave well when encountering complex reasoning scenarios.}
\label{fig:vis}
\end{center}
\end{figure*}

\begin{table*}[ht]
\small
\centering
\vspace{-0.4cm}
\setlength{\abovecaptionskip}{0cm}
\renewcommand\arraystretch{1.1}
\caption{Examples of transferring QA pairs to declarative sentences on the SUTD-TrafficQA dataset. Different traffic-related reasoning tasks are included.}
\scalebox{0.9}
{%
\begin{tabular}{@{}ll@{}}
\bottomrule
\midrule
\multicolumn{1}{l|}{\textbf{Task}} & Basic Understanding \\ \midrule
 \multicolumn{1}{l|}{\textbf{Question}} & Which area has been damaged on the vehicle being hit?\\ \midrule
 \multicolumn{1}{l|}{\textbf{Answers}} & \begin{tabular}[c]{@{}l@{}}
 Back \qquad  Front   \qquad  Side \\
 \end{tabular} \\ \midrule
 \multicolumn{1}{l|}{\textbf{Declarative sentences}} & \begin{tabular}[c]{@{}l@{}}
 Back has been damaged on the vehicle being hit. \\
 Side has been damaged on the vehicle being hit. \\
 Front has been damaged on the vehicle being hit. \\
 \end{tabular} \\ \midrule
 \midrule

\multicolumn{1}{l|}{\textbf{Task}} & Attribution\\ \midrule
 \multicolumn{1}{l|}{\textbf{Question}} & What could possibly cause this accident? \\ \midrule
 \multicolumn{1}{l|}{\textbf{Answers}} & \begin{tabular}[c]{@{}l@{}}
Obstructed by unexpected objects \\
Sudden braking of a vehicle \\
Violation of traffic rules by pedestrians \\
Sudden or extreme movement by a vehicle \\
 \end{tabular} \\ \midrule
 \multicolumn{1}{l|}{\textbf{Declarative sentences}} & \begin{tabular}[c]{@{}l@{}}
Obstructed by unexpected objects could possibly cause this accident. \\
Sudden braking of a vehicle could possibly cause this accident. \\
Violation of traffic rules by pedestrians could possibly cause this accident. \\ Sudden or extreme movement by a vehicle could possibly cause this accident. \\
 \end{tabular} \\ \midrule
\midrule

\multicolumn{1}{l|}{\textbf{Task}} & Introspection \\ \midrule
 \multicolumn{1}{l|}{\textbf{Question}} & Can this road infrastructure prevent head-on collision? \\ \midrule
 \multicolumn{1}{l|}{\textbf{Answers}} & \begin{tabular}[c]{@{}l@{}}
No, the road is unmarked \\
Yes, the divider between two directions is marked clearly \\
 \end{tabular} \\ \midrule
 \multicolumn{1}{l|}{\textbf{Declarative sentences}} & \begin{tabular}[c]{@{}l@{}}
This road infrastructure cannot prevent head-on collision, the road is unmarked. \\
This road infrastructure can prevent head-on collision, the divider between two directions is marked clearly. \\
 \end{tabular} \\ \midrule
\midrule

\multicolumn{1}{l|}{\textbf{Task}} & Counterfactual Inference\\ \midrule
 \multicolumn{1}{l|}{\textbf{Question}} & Would the accident still occur if the driver slows down in time? \\ \midrule
 \multicolumn{1}{l|}{\textbf{Answers}} & \begin{tabular}[c]{@{}l@{}}
Yes     \qquad  No \\
 \end{tabular} \\ \midrule
 \multicolumn{1}{l|}{\textbf{Declarative sentences}} & \begin{tabular}[c]{@{}l@{}}
The accident still occur if the driver slows down in time. \\
The accident would not occur if the driver slows down in time.
 \end{tabular} \\ \midrule
\midrule

\multicolumn{1}{l|}{\textbf{Task}} & Reverse Reasoning\\ \midrule
 \multicolumn{1}{l|}{\textbf{Question}} & Which could be the reason for this accident? \\ \midrule
 \multicolumn{1}{l|}{\textbf{Answers}} & \begin{tabular}[c]{@{}l@{}}
Traffic light violation \\
Retrograde vehicles \\
Improper lane change \\
Obstructed view or limited visibility \\
 \end{tabular} \\ \midrule
 \multicolumn{1}{l|}{\textbf{Declarative sentences}} & \begin{tabular}[c]{@{}l@{}}
Traffic light violation could be the reason for this accident. \\
Retrograde vehicles could be the reason for this accident. \\
Improper lane change could be the reason for this accident. \\
Obstructed view or limited visibility could be the reason for this accident. \\
 \end{tabular} \\ \midrule
\midrule

\multicolumn{1}{l|}{\textbf{Task}} & Event Forecasting \\ \midrule
 \multicolumn{1}{l|}{\textbf{Question}} & How much damage will the vehicle(s) receive after collision? \\ \midrule
 \multicolumn{1}{l|}{\textbf{Answers}} & \begin{tabular}[c]{@{}l@{}}
Nearly no damage \\
Significant deformation \\
Some scratches \\
\end{tabular} \\ \midrule
 \multicolumn{1}{l|}{\textbf{Declarative sentences}} & \begin{tabular}[c]{@{}l@{}}
 The vehicle (s) will receive significant deformation after collision. \\
 The vehicle (s) will receive nearly no damage after collision. \\
 The vehicle (s) will receive some scratches after collision. \\
 \end{tabular} \\ \midrule
\bottomrule
\end{tabular}%
}
\label{tab:template-generation}
\vspace{-0.4cm}
\end{table*}

\begin{table*}[ht]
\small
\centering
\vspace{-0.4cm}
\setlength{\abovecaptionskip}{0cm}
\renewcommand\arraystretch{0.95}
\caption{Examples of transferring QA pairs to declarative sentences. Different type of questions are included.}
\scalebox{0.9}
{%
\begin{tabular}{@{}ll@{}}
\bottomrule
\midrule
\multicolumn{1}{l|}{\textbf{Question Type}} & Where \\ \midrule
 \multicolumn{1}{l|}{\textbf{Question}} & Where was the video taken?  \\ \midrule
 \multicolumn{1}{l|}{\textbf{Answers}} & \begin{tabular}[c]{@{}l@{}}
 A crossroad \\ 
 The countryside \\
 Road in the city \\
 Forest \\
 \end{tabular} \\ \midrule
 \multicolumn{1}{l|}{\textbf{Declarative sentences}} & \begin{tabular}[c]{@{}l@{}}
The video was taken in a crossroad. \\
The video was taken in the countryside. \\
The video was taken in the city. \\
The video was taken in Forest. \\ 
 \end{tabular} \\ \midrule
 \midrule

\multicolumn{1}{l|}{\textbf{Question Type}} & Why \\ \midrule
 \multicolumn{1}{l|}{\textbf{Question}} & Why did the accident occur when the road is clear? \\ \midrule
 \multicolumn{1}{l|}{\textbf{Answers}} & \begin{tabular}[c]{@{}l@{}}
Vehicle malfunction. \\
Trying to avoid something on the road. \\
Driver was not paying attention to the road. \\
Uneven road, full of potholes. \\
\end{tabular} \\ \midrule
 \multicolumn{1}{l|}{\textbf{Declarative sentences}} & \begin{tabular}[c]{@{}l@{}}
The accident occurred when the road is clear because of vehicle malfunction. \\
The accident occurred when the road is clear because of trying to avoid something on the road. \\
The accident occurred when the road is clear because driver was not paying attention to the road. \\
The accident occurred when the road is clear because of uneven road, full of potholes. \\
 \end{tabular} \\ \midrule

\multicolumn{1}{l|}{\textbf{Question Type}} & How \\ \midrule
 \multicolumn{1}{l|}{\textbf{Question}} & How did the truck get involved in the accident? \\ \midrule
 \multicolumn{1}{l|}{\textbf{Answers}} & \begin{tabular}[c]{@{}l@{}}
The truck is being hit from behind. \\
The truck is being hit from the side. \\
\end{tabular} \\ \midrule
 \multicolumn{1}{l|}{\textbf{Declarative sentences}} & \begin{tabular}[c]{@{}l@{}}
The truck get involved in the accident by being hit from behind. \\
The truck get involved in the accident by being hit from the side. \\
 \end{tabular} \\ \midrule

\multicolumn{1}{l|}{\textbf{Question Type}} & How many \\ \midrule
 \multicolumn{1}{l|}{\textbf{Question}} & How many lanes does the road have in single direction? \\ \midrule
 \multicolumn{1}{l|}{\textbf{Answers}} & \begin{tabular}[c]{@{}l@{}}
Two, ~ Only one, ~ Three to five  
\end{tabular} \\ \midrule
 \multicolumn{1}{l|}{\textbf{Declarative sentences}} & \begin{tabular}[c]{@{}l@{}}
The road has two in single direction. \\ 
The road has only one in single direction. \\
The road has three to five in single direction. \\
 \end{tabular} \\ \midrule

\multicolumn{1}{l|}{\textbf{Question Type}} & What's \\ \midrule
 \multicolumn{1}{l|}{\textbf{Question}} & What's the condition of the road surface? \\ \midrule
 \multicolumn{1}{l|}{\textbf{Answers}} & \begin{tabular}[c]{@{}l@{}}
The road is wet. \\
The road is covered by snow and ice.\\
The road is smooth and clean. \\
The road is dusty or muddy. \\
\end{tabular} \\ \midrule
 \multicolumn{1}{l|}{\textbf{Declarative sentences}} & \begin{tabular}[c]{@{}l@{}}
The condition of the road surface is wet. \\ The condition of the road surface is covered by snow and ice. \\
The condition of the road surface is smooth and clean. \\
The condition of the road surface is dusty or muddy. \\
 \end{tabular} \\ \midrule

\multicolumn{1}{l|}{\textbf{Question Type}} & General Question \\ \midrule
 \multicolumn{1}{l|}{\textbf{Question}} & Are there any trees along the road? \\ \midrule
 \multicolumn{1}{l|}{\textbf{Answers}} & \begin{tabular}[c]{@{}l@{}}
Yes, ~No
\end{tabular} \\ \midrule
 \multicolumn{1}{l|}{\textbf{Declarative sentences}} & \begin{tabular}[c]{@{}l@{}}
There are some trees along the road. \\ There are not any trees along the road. \\
 \end{tabular} \\ \midrule

\multicolumn{1}{l|}{\textbf{Question Type}} & General Question \\ \midrule
 \multicolumn{1}{l|}{\textbf{Question}} & Did a car violate the traffic light? \\ \midrule
 \multicolumn{1}{l|}{\textbf{Answers}} & \begin{tabular}[c]{@{}l@{}}
Yes, ~No
\end{tabular} \\ \midrule
 \multicolumn{1}{l|}{\textbf{Declarative sentences}} & \begin{tabular}[c]{@{}l@{}}
A car violated the traffic light. \\
A car did not violate the traffic light. \\
 \end{tabular} \\ \midrule

\bottomrule
\end{tabular}%
}

\label{tab:template-generation-questiontype}
\vspace{-0.4cm}
\end{table*}

\end{document}